\documentclass[letterpaper, 10 pt, conference]{ieeeconf}  
\IEEEoverridecommandlockouts                  
\usepackage[letterpaper, left=0.75in, right=0.75in, bottom=0.75in, top=0.75in]{geometry}
\usepackage{graphics} 

\usepackage{amsmath}  
\usepackage{amssymb}  
\usepackage[export]{adjustbox} 
\usepackage{bm}
\usepackage{balance}
\usepackage[caption=false,font=normalsize
]{subfig}
\usepackage{cuted}
\usepackage{capt-of}
\usepackage{notoccite}
\usepackage{CJKutf8}
\usepackage{tikz}
\usepackage[compact]{titlesec}
\usepackage{stfloats} 
\usepackage{setspace} 
\usepackage{ulem} 

\usepackage[linesnumbered,ruled,vlined]{algorithm2e}
\SetKwInput{KwIn}{Input}
\SetKwInput{KwOut}{Output}
\usepackage{booktabs}
\usepackage{makecell}
\usepackage{multirow}
\usepackage{graphicx}
\usepackage{pifont}
\usepackage{mathtools}
\usepackage{float}
\usepackage{tabularx}
\usepackage{xcolor}
\let\labelindent\relax
\usepackage{enumitem}
\newcommand{\Tobs}{T_{\text{obs}}}
\newcommand{\Tpred}{T_{\text{pred}}}
\newcommand{\cmark}{\ding{51}}%
\newcommand{\xmark}{\ding{55}}%
\newcommand{\vx}{\mathbf{x}}      
\newcommand{\vX}{\mathbf{X}}      
\newcommand{\vY}{\mathbf{Y}}      

\newcommand{\etal}{\textit{et al.~}}

\newcolumntype{D}{c}                 
\newcolumntype{O}{@{\hskip -6pt}c}

\makeatletter
 \let\NAT@parse\undefined
 \makeatother
\usepackage[pdftex,
        colorlinks=true,
        bookmarks=true,
        citecolor=red,
        linkcolor=blue,
        pagebackref=true,
        pdfstartview=Fit,
        breaklinks=true
            ]{hyperref}
\usepackage{cleveref}

\title{\LARGE\bf
LLM-Grounded Dynamic Task Planning with Hierarchical Temporal Logic for Human-Aware Multi-Robot Handover
\vspace{-0.5cm}
}
\author{Shuyuan Hu$^{1\dagger}$, Tao Lin$^{1,2\dagger}$, Kai Ye$^{1,3}$, Tianwei Zhang$^{1,3*}$%
\thanks{$^{\dagger}$These authors contributed equally to this work.}%
\thanks{$^{1}$The Shenzhen Institute of Artificial Intelligence and Robotics for Society, Shenzhen, China}%
\thanks{$^{2}$Harbin Institute of Technology, Harbin, China}%
\thanks{$^{3}$The Chinese University of Hong Kong-Shenzhen, Shenzhen, China}%
\thanks{* Corresponding Author: \tt\small {zhangtianwei@cuhk.edu.cn}}
\thanks{This work was supported by the Shenzhen Science and Technology Program (Grant No. JSGGKQTD20221101115656029, ZDCY20250901094531003 and KJZD20230923113801004)}
}
\begin{document}
\begin{CJK}{UTF8}{gbsn}

\thispagestyle{empty}
\pagestyle{empty}
\maketitle
\begin{strip}
\centering
\begin{minipage}{\textwidth}
\vspace{-2.8cm}
  \includegraphics[width=\textwidth]{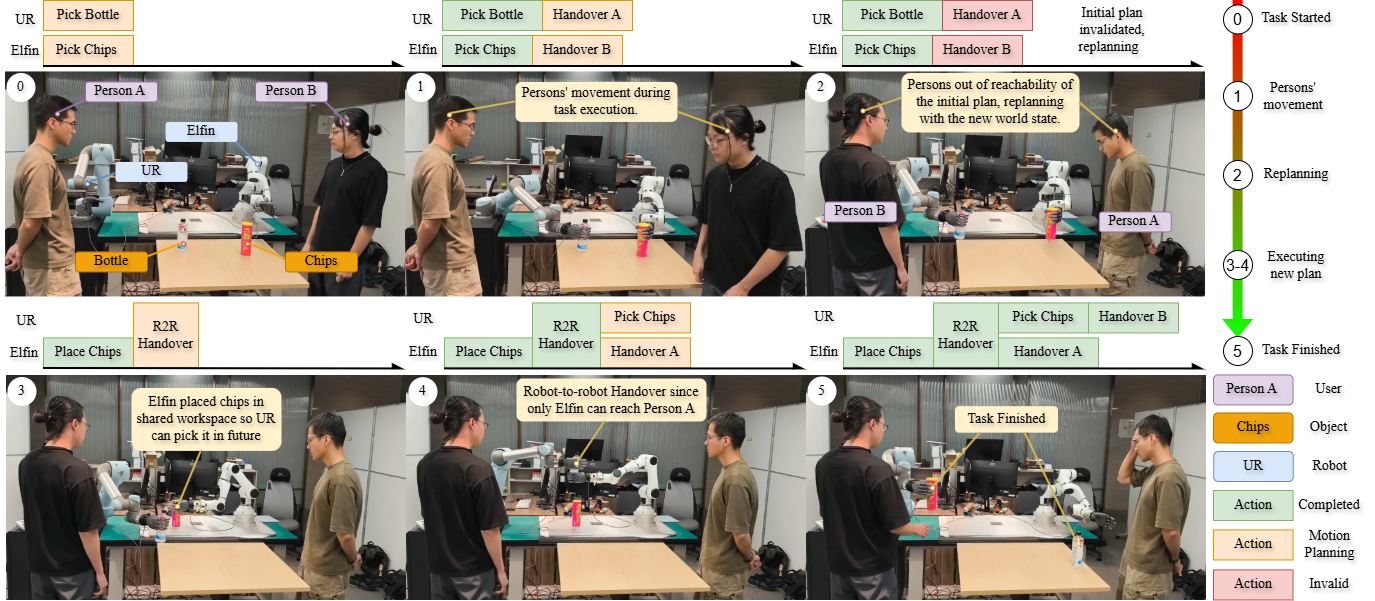}
  \captionof{figure}{A demonstration of the proposed dynamic planning with the human instruction, \textbf{``Give Person A the bottle and Person B the chips."} At Snapshot 1, perception detects user motion and starts predictive replanning in the background. At Snapshot 2, the updated user positions violate the current handover precondition, so execution is halted and the system switches to the replanned suffix once it is available.}
  \label{fig:overview_of_system}
\vspace{-15pt}
\end{minipage}
\end{strip}
\begin{abstract}
Large Language Models (LLMs) enable non-experts to specify open-world multi-robot tasks, but the generated plans are often kinematically infeasible and inefficient in long-horizon settings. Formal methods such as Linear Temporal Logic (LTL) offer correctness and optimality guarantees, yet they are typically offline and scale poorly. To bridge this gap, we propose a neuro-symbolic framework that grounds human instructions into hierarchical LTL$_f$ specifications (i.e., LTL on finite traces) and solves the resulting Simultaneous Task Allocation and Planning (STAP) problem. Unlike static approaches, our system handles stochastic environmental changes—such as user motion or updated instructions—through a receding-horizon planning (RHP) loop with real-time perception, dynamically refining plans over a hierarchical state space. Experiments in simulation and on real robots demonstrate that our approach significantly outperforms baseline methods in success rate and interaction fluency while reducing replanning overhead.
\end{abstract}

\section{Introduction}

Long-horizon multi-robot collaboration in dynamic, human-centric environments requires robust coordination under changing human motion and task updates \cite{caai}. We use \textit{long-horizon} to refer to tasks composed of multiple temporally constrained subtasks whose feasible allocation and execution sequence may change during execution. Two prominent but largely separate paradigms have emerged to address this challenge. On one hand, LLM- and Vision-Language-Model (VLM)-based planners demonstrate remarkable flexibility in interpreting human intent and proposing plans in open-world environments \cite{dovsg2024,zhu2025dexterllmdynamicexplainablecoordination}. Yet, they often lack formal guarantees, struggling to ensure that generated plans are dynamically feasible and logically sound, especially in multi-agent settings. On the other hand, \textit{formal methods} such as LTL \cite{luo2025simultaneous} provide a rigorous mathematical framework for specifying complex tasks and synthesising plans with provable correctness. 
Their primary limitation, however, is their reliance on a static, fully known world model, making them brittle and unsuitable for direct application in dynamic environments \cite{my-survey-bir}.
Moreover, integrating formal planning methods into a real-time system poses a significant challenge: the time complexity of mainstream formal methods increases exponentially with the state space \cite{kurtz2023temporal}.

To bridge this gap, we propose a framework that grounds LLM reasoning into a hierarchical variant of LTL$_{f}$, denoted H-LTL$_{f}$, which significantly reduces the search space of multi-robot planning. By operating within a receding-horizon loop, our system synthesizes high-level strategy with real-time perception, enabling robust execution in dynamic, human-centric environments, as shown in Fig.~\ref{fig:overview_of_system}: Snapshot~1 triggers predictive replanning, and Snapshot~2 triggers safety halt and suffix switching.
 
Our main contributions are:
\begin{itemize}[leftmargin=*, topsep=0pt, itemsep=1pt, parsep=0pt, partopsep=0pt]
    \item We propose a neuro-symbolic method for online dynamic multi-robot planning of long-horizon tasks in human-aware environments.
    \item We extend H-LTL$_{f}$ planning from single-robot leaf execution to strongly coupled cooperative STAP via a coalition-aware unified graph.
    \item We implement and validate our framework on multi-robot handover tasks in dynamic human-aware environments.
\end{itemize}


\section{Related Works}\label{char2}

\subsection{Formal Methods for Planning}
Formal planning methods provide a mathematically rigorous way to specify and synthesise robot behaviour. Temporal-logic formalisms such as LTL can express rich temporal requirements over task sequences and safety conditions; when combined with automata planners, they offer completeness and, in many cases, optimality guarantees \cite{capacitor}. 
Building on this foundation, recent LLM-based approaches increasingly integrate with formal planners rather than replacing them. LaMMA-P \cite{zhang2025lamma} couples LLM-driven subtask allocation and PDDL problem generation with a classical planner (Fast Downward) for long-horizon multi-agent tasks, while DEXTER-LLM \cite{zhu2025dexterllmdynamicexplainablecoordination} integrates LTL-based mission abstraction, LLM subtask generation, and optimisation-based scheduling for dynamic multi-robot coordination.

However, traditional temporal-logic planning methods struggle with long-horizon tasks where all requirements are encoded as a single flat LTL formula: the corresponding automaton quickly becomes intractable and difficult to interpret as task complexity and horizon grow. Luo et al.\ introduce hierarchical LTL specifications \cite{luo2025simultaneous} that decompose a global specification into loosely coupled sub-specifications, significantly reducing automaton size and scaling formal guarantees to long-horizon multi-robot tasks. Luo \etal\ further provide the formal backbone for Nl2HLTL2Plan \cite{xu2024nl2hltl2plan}. Nl2HLTL2Plan uses an LLM to translate natural-language instructions into hierarchical LTL specifications, but it solves the resulting tasks in an open-loop manner without online feedback.

\subsection{Language-Conditioned Robotic Planning}
LLM- and VLM-based methods have recently shown promise as zero-shot planners in robotics, decomposing long-term natural-language goals into ordered subtask action sequences for a single robot.
While powerful, these systems predominantly target single-robot settings; two critical gaps remain when scaling LLM-based planning paradigms from single-robot autonomy to truly collaborative teams: (i) task allocation across multiple agents under spatio-temporal constraints and (ii) retaining robustness when objects, layouts, or partner robots change at run time. COHERENT~\cite{coherent2024} extends LLM-driven planning to heterogeneous multi-robot teams while modelling it as a static sequential planning problem without spatio-temporal coordination. 

\begin{table}[tbhp]
\small
\centering
\caption{ Comparison of LLM-based multi-robot planning methods}
\label{tab:methods_comparison}
\vspace{-5pt}
\resizebox{\linewidth}{!}{%
\begin{tabular}{lccc@{\hspace{2pt}}c}
\toprule
    & \makecell{Formal \\ Planning} 
    & \makecell{Dynamic \\ Planning}  
    & \makecell{Dynamic \\ Contexts}  
    & \makecell{Real-World \\ Deployment} \\ 
\midrule
\cite{kannan2023smart}
& \xmark         & \xmark               & \xmark                           & \xmark \\
\cite{wang2025dartllmdependencyawaremultirobottask}  
& \xmark         & Offline  & \makecell[c]{O\mbox{-}MOVE}      & \cmark \\
\cite{coherent2024} 
& \xmark         & Offline  & \makecell[c]{FAIL}       & \xmark \\
\cite{mandi2024roco}
& \xmark         & Offline    & \makecell[c]{RES, FAIL}        & \cmark \\
\cite{xu2024nl2hltl2plan}      
& H-LTL$_{f}$ & \xmark               & \xmark                           & \cmark \\
\cite{zhang2025lamma}               
& PDDL           & Offline    & \makecell[c]{FAIL}               & \xmark \\
\cite{zhu2025dexterllmdynamicexplainablecoordination} 
& TL &  Online & \makecell[c]{O\mbox{-}MOVE, RES\\GOAL, FAIL} & \xmark \\
\midrule
\textbf{\textsc{Ours}}                      
& H-LTL$_{f}$ & Online          & \makecell[c]{H\mbox{-}MOVE, O\mbox{-}MOVE,\\ RES, GOAL, FAIL}  & \cmark \\
\bottomrule
\end{tabular}%
}
\parbox{\linewidth}{Abbrev.: \textbf{H-MOVE} = human position/posture change, \textbf{O-MOVE} = object moved/pose changed, \textbf{GOAL} = instruction change or priority update, \textbf{RES} = multi-robot resource conflict or deadlock, \textbf{FAIL} = execution-level failure such as grasp/control failure, \textbf{PDDL} = Planning Domain Definition Language; \cmark\ = supported; \xmark\ = not supported or not reported.}
\vspace{-5pt}
\end{table}

\subsection{Human-Aware Multi-Robot Collaboration}
Foundational tasks in robot-human and multi-robot teams \cite{hrc}, such as object handovers and co-manipulation \cite{mascaro2023humanintention}, have been studied thoroughly. Such works, however, typically handle short, structured interactions (e.g., passing an item) rather than long-horizon missions. 
Emerging works leverage foundation models for more generalizable human-robot teaming by allowing humans to provide high-level guidance or corrections~\cite{mandi2024roco}. However, most prior approaches rely on open-loop planning, which involves human dynamics only at discrete checkpoints (e.g., task initiation or error correction) and assumes a relatively \textbf{static environment} during execution \cite{wang2025dartllmdependencyawaremultirobottask}. While recent work such as DEXTER-LLM~\cite{zhu2025dexterllmdynamicexplainablecoordination} has begun to explore online coordination mechanisms to handle dynamic updates, its validation remains confined to simulation environments. As summarised in Table~\ref{tab:methods_comparison}, our approach is among the first to deploy an LLM-based online planning framework for dynamic multi-robot collaboration in the real world.

\begin{figure*}[tbhp]
\centering
\includegraphics[width=0.9\textwidth]{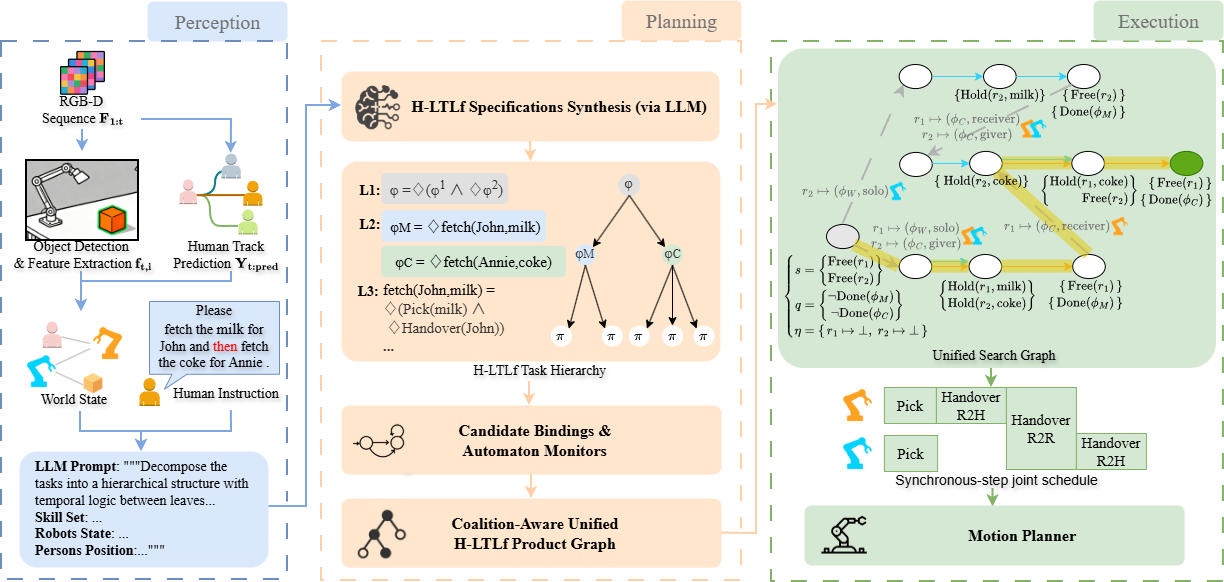}
\vspace{-5pt}
\caption{The proposed method flowchart. In the \textit{Unified Search Graph} panel, we illustrate planning for two leaf specifications and two robots using the unified state $x=(s,\eta,q)$, where $s$ is the joint robot state, $\eta$ is the role-aware binding map, and $q$ stores automaton-monitor states (one-time leaf completion). \textcolor{gray}{Gray dashed arrows} denote zero-cost bind/unbind switches that update $\eta$. Solid arrows denote skill-execution steps:  \textcolor[RGB]{252,162,45}{orange} for $r_1$ and \textcolor[RGB]{0,143,176}{blue} for $r_2$; synchronous multi-robot steps (e.g., handover) overlay both colors on the same edge. For clarity, only  \textcolor{gray}{the initial node} shows full $(s,\eta,q)$; later nodes record only state deltas. The path marked in \textcolor{yellow}{yellow} highlights the selected plan.}
\label{fig:architecture}
\vspace{-20pt}
\end{figure*}
\section{\textsc{System Framework And Methods}}
We formulate the collaboration problem as STAP in dynamic environments. In this work, we instantiate the available atomic skills as \textit{Pick}, \textit{Place}, and \textit{Handover}; the H-LTL$_f$ representation and receding-horizon planner are skill-agnostic, while the fixed RGB-D camera and cylindrical objects are validation-specific choices.
Our method introduces a closed-loop task-planning framework that empowers multi-robot systems to execute long-horizon tasks from natural-language instructions in dynamic, human-aware environments. The architecture, depicted in Fig.~\ref{fig:architecture}, comprises three core, interconnected stages: (1) a real-time, open-vocabulary 3D perception module that builds and maintains a semantic representation of the world; (2) an LLM grounds the instruction into an H-LTL$_f$ hierarchy, after which the planner compiles monitors and constructs a unified search graph for on-the-fly optimal search; and (3) a low-level robot dispatch and control module dynamically selects and executes the next action from the unified search graph.

\subsection{Open-vocabulary 3D Perception}
We use a fixed RGB-D camera. For each incoming frame $F_t$, we build an object-centric 3D scene to track objects and predict human trajectories in parallel.

\textbf{Open-vocabulary 2D parsing.}
We first apply Recognize-Anything \cite{zhang2024recognize} to obtain open-vocabulary class labels
$\{c_{t,i}\}_{i=1}^{N_o}$ for all $N_o$ objects detected.
These labels condition Grounding~DINO \cite{liu2024grounding} to generate 2D bounding boxes
$\{b_{t,i}\}$, which are refined into pixel-accurate masks
$\{m_{t,i}\}$ by SAM 2 \cite{ravi2024sam}.
From each $(b_{t,i}, m_{t,i})$, we extract two images (a crop and a background-removed mask) and compute CLIP-based \cite{radford2021learning} visual features, fused by a weighted sum to form a single descriptor:
\begin{equation}
    f_{t,i} \coloneqq \mathrm{Embed}\!\left(F_t^{\text{rgb}},\, b_{t,i},\, m_{t,i}\right).
\end{equation}



\textbf{Language-Guided 3D Localization.}
To locate the target object specified in the human instruction, we compute the cosine similarity between the CLIP text embedding of the target object name and the visual feature embeddings $f_{t,i}$ of all detected objects. The object with the highest similarity score is identified as the target. Subsequently, the target object’s mask $m_{t,i}$ is back-projected with its depth to produce the object point cloud $P_{t,i}$ for downstream pickup planning.

\textbf{Person tracking and trajectory prediction.} In parallel, we run YOLOv11 for person detection and a lightweight, self-trained person classifier to maintain identities in image space. A compact recurrent network forecasts short-horizon human trajectories, providing future human position estimates for downstream planning. For each person $i\in\{1,\dots,N_h\}$, the observed (3D) trajectory over $\Tobs$ frames is
\begin{equation}
  \vX^{(i)}_{1:\Tobs}
  = \{\vx^{(i)}_1,\vx^{(i)}_2,\dots,\vx^{(i)}_{\Tobs}\}, \quad \vx^{(i)}_t \in \mathbb{R}^3.
\end{equation}
and our prediction is denoted by
\begin{equation}
  \vY^{(i)}_{1:\Tpred}
  = \{\hat{\vx}^{(i)}_{\Tobs+1},\dots,\hat{\vx}^{(i)}_{\Tobs+\Tpred}\}.
\end{equation}

\textbf{Instruction grounding to formal specs.}
Given the perceived world state $w_t$ (semantic scene graph with tracked entities) and the human instruction, we query an LLM to produce a grounded H-LTL$_f$ specification $\Phi$ (and referenced entities) that is consistent with the current scene graph; $\Phi$ is then used by the downstream hierarchical planning module.

\subsection{Hierarchical LTL\texorpdfstring{$_f$}{f}}
We adopt H-LTL$_f$ from~\cite{luo2025simultaneous}. An H-LTL$_f$ specification
$\Phi=\{\phi_k^i\}$ has $K$ levels and satisfies~\cite[Def.~4.1--4.3]{luo2025simultaneous};
satisfaction follows~\cite[Def.~4.4--4.6]{luo2025simultaneous}. Let $\Phi_{\text{leaf}}$ be the leaf set.

\subsection{Coalition-Aware Unified H-LTL$_f$ Product Graph}\label{subsec:team_models}
We build a unified search graph realizing the H-LTL$_f$ bottom-up semantics~\cite{luo2025simultaneous}, enabling STAP for
single-robot leaves and 2-robot handover leaves with makespan cost under synchronous execution. The graph jointly searches allocation, role binding, and execution.

\subsubsection{Robot Team Model and Skill Library}
Let $\mathcal{R}\triangleq\{1,\dots,N\}$. Each robot $r\in\mathcal{R}$ is a weighted transition system
$T_r=(S_r,s_r^0,\rightarrow_r,\mathcal{AP},L_r,c_r)$; the joint state is
$\mathbf{s}=(s_1,\dots,s_N)\in\mathbf{S}\triangleq\prod_{r=1}^N S_r$.

\textbf{Human-aware context.}
We use a discrete context label $\ell_t\in\mathcal{L}$ (e.g., the current human workspace index over regions
$\{\mathcal{W}_j\}_{j=1}^{N_{\mathrm{ws}}}$) to capture time-varying reachability/safety constraints induced by human motion. 
The mapping $\ell_t=\ell(w_t)$ from the perceived world state $w_t$ is defined in Sec.~\ref{subsec:dynamic_execution}.

Let $\mathcal{J}$ be the skill library; each $a\in\mathcal{J}$ has executors $\mathrm{Exec}(a)\subseteq\mathcal{R}$,
precondition $\mathrm{Pre}(\mathbf{s},\ell_t,a)$, post map $\mathbf{s}'=\mathrm{Post}(\mathbf{s},a)$, and duration $d(a)\ge 0$. For a synchronous set $U$ with disjoint executors, define $\mathrm{Post}(\mathbf{s},U)$ as the parallel composition of skill posts:
$\mathrm{Post}(\mathbf{s},U)\triangleq \mathrm{Post}(\cdots \mathrm{Post}(\mathrm{Post}(\mathbf{s},a_1),a_2)\cdots,a_{|U|})$,
where the order is irrelevant due to disjoint executors.

\textbf{Synchronous step.}
At step $t$, choose $U_t\subseteq\mathcal{J}$ with disjoint executors:
\begin{equation}
\forall a\neq b\in U_t:\ \mathrm{Exec}(a)\cap\mathrm{Exec}(b)=\emptyset.
\label{eq:disjoint_exec_new}
\end{equation}
If $\mathrm{Pre}(\mathbf{s}_t,\ell_t,a)$ holds for all $a\in U_t$, then
\begin{equation}
\mathbf{s}_{t+1}=\mathrm{Post}(\mathbf{s}_t,U_t),
\label{eq:post_joint}
\end{equation}
and the step duration is the parallel makespan
\begin{equation}
\Delta T_t=\max_{a\in U_t} d(a),\quad \max\emptyset=0.
\label{eq:makespan_step_new}
\end{equation}
Thus $J(\pi)=\sum_{t=0}^{T-1}\Delta T_t$.

\textbf{Atomic handover.}
$\mathrm{Handover}(o;r_g,r_r)$ is a 2-robot skill with $\mathrm{Exec}(a)=\{r_g,r_r\}$ and
$\mathrm{Pre}(\mathbf{s},\ell_t,a)\equiv \mathrm{Hold}(r_g,o)\wedge \mathrm{Free}(r_r)\wedge \mathrm{Rendezvous}(r_g,r_r,o;\ell_t)$.
Its postcondition swaps possession: $\mathrm{Hold}(r_g,o)\mapsto \mathrm{Free}(r_g)$ and
$\mathrm{Free}(r_r)\mapsto \mathrm{Hold}(r_r,o)$.

\subsubsection{Leaf Types, Coalition Selection, and Assignment}
Partition $\Phi_{\text{leaf}}=\Phi^{(1)}\ \dot{\cup}\ \Phi^{(2)}$ into single-robot and 2-robot (handover) leaves; each leaf completes once.
Robots are not preassigned: for object $o$ and target $u$, define feasible sets
$\mathcal{R}_{\mathrm{pick}}(o)$ and $\mathcal{R}_{\mathrm{int}}(u)$. For $\phi=\phi_{\mathrm{ho}}(o,u)$,
candidate bindings are
\begin{equation}
\mathcal{B}_\phi \triangleq \{(r_g,r_r)\in \mathcal{R}\times\mathcal{R}\mid r_g\in\mathcal{R}_{\mathrm{pick}}(o),\ r_r\in\mathcal{R}_{\mathrm{int}}(u)\}.
\label{eq:candidate_binding_new}
\end{equation}

\textbf{Role-aware assignment.}
This is needed because handovers have asymmetric giver/receiver roles. Maintain $\eta:\mathcal{R}\rightarrow (\Phi_{\text{leaf}}\times\mathcal{K})\cup\{\bot\}$ with
$\mathcal{K}=\{\textsf{solo},\textsf{giver},\textsf{receiver}\}$, and $\eta(r)=\bot$ meaning idle.
For a leaf $\phi$, let $C_\phi(\eta)\triangleq\{r\mid \eta(r)=(\phi,k)\}$.
Feasibility is
\begin{gather}
C_\phi(\eta)=\emptyset\ \ \text{or}\ \notag\\
\begin{cases}
C_\phi(\eta)=\{r\}\wedge \eta(r)=(\phi,\textsf{solo}), & \phi\in\Phi^{(1)},\\
C_\phi(\eta)=\{r_g,r_r\}\wedge \eta(r_g)=(\phi,\textsf{giver})\\ \wedge \eta(r_r)=(\phi,\textsf{receiver})\wedge (r_g,r_r)\in\mathcal{B}_\phi, & \phi\in\Phi^{(2)}.
\end{cases}
\label{eq:eta_feas}
\end{gather}
Between steps, $\eta$ may change via zero-time bind/unbind.

\subsubsection{H-LTL$_f$ Monitors with One-Time Leaf Completion}
Let $\mathcal{T}_h=(V_h,E_h)$ be the hierarchy tree and $\mathrm{Ch}(\varphi)$ the children set. Each node $\varphi\in V_h$
is compiled into an NFA $\mathcal{A}_\varphi=(Q_\varphi,Q_\varphi^0,\Sigma_\varphi,\delta_\varphi,F_\varphi)$ and determinized
(or tracked as subset-states), so $q_\varphi(t)$ is well-defined.

\textbf{One-shot leaves.} Leaf monitors are one-shot: once $q_\phi$ first reaches $F_\phi$, it becomes absorbing, so a completed subtask contributes only once.

\textbf{Leaf propositions and events.}
For each leaf $\phi\in\Phi_{\text{leaf}}$, let $\mathcal{AP}_\phi\subseteq\mathcal{AP}$ be its relevant propositions and define
$L_r^\phi(s_r,\eta(r))\triangleq L_r(s_r)\cap\mathcal{AP}_\phi$.
Let $L^{\mathrm{rel}}_\phi(\mathbf{s},\eta,\ell_t)\subseteq\mathcal{AP}_\phi$ denote relational and human-aware predicates under context $\ell_t$
(e.g., possession/rendezvous and region-safety constraints).
Define the event map $E(U_t)\triangleq\{e(a)\in\mathcal{AP}\mid a\in U_t\}$.

In particular, for a handover instance $a=\mathrm{Handover}(o;r_g,r_r)$ assigned to leaf $\phi=\phi_{\mathrm{ho}}(o,u)$ by $\eta_t$, we set $e(a)=\mathsf{HO}_\phi$.

For each leaf $\phi$, the consumed symbol at step $t$ is
\begin{gather}
\sigma_\phi(\mathbf{s}_{t+1},\ell_t,\eta_t,U_t)
=\Big(\!\bigcup_{r\in C_\phi(\eta_t)}\! L^\phi_r(s_{t+1,r},\eta_t(r))\Big)\ \cup\ \notag\\
L^{\mathrm{rel}}_\phi(\mathbf{s}_{t+1},\eta_t,\ell_t) 
\cup\ E(U_t).
\label{eq:leaf_symbol_new}
\end{gather}
For handover leaves, use $\phi\equiv\Diamond\,\mathsf{HO}_\phi$ and
\begin{equation}
\mathsf{HO}_\phi\in E(U_t)\ \Leftrightarrow\ \mathrm{Handover}(o;r_g,r_r)\in U_t,
\label{eq:handover_leaf_new}
\end{equation}
where $(r_g,r_r)=C_\phi(\eta_t)$ and $\phi=\phi_{\mathrm{ho}}(o,u)$.
Define $\mathrm{DoneNow}(\xi,t)=1 \Leftrightarrow (q_\xi(t)\notin F_\xi)\wedge(q_\xi(t+1)\in F_\xi)$ for any hierarchy node $\xi\in V_h$,
\begin{equation}
\sigma_\varphi(t)=\{\xi\in \mathrm{Ch}(\varphi)\mid \mathrm{DoneNow}(\xi,t)=1\}.
\label{eq:parent_symbol_new}
\end{equation}
Monitors update bottom-up.

\subsubsection{Unified Search Graph and Objective}
A search node is $x_t=(\mathbf{s}_t,\eta_t,\mathbf{q}_t)$ with $\mathbf{q}_t=(q_\varphi(t))_{\varphi\in V_h}$.

Let $\Pi(x\rightsquigarrow\mathcal{F})$ denote the set of feasible paths in the unified search graph from $x$ to the accepting set $\mathcal{F}$.

A transition applies (i) optional zero-time $\eta_t\mapsto\eta_t^+$ satisfying~\eqref{eq:eta_feas},
(ii) feasible $U_t$ under $\mathrm{Pre}(\mathbf{s}_t,\ell_t,\cdot)$ and disjoint executors~\eqref{eq:disjoint_exec_new},
and (iii) monitor updates via \eqref{eq:leaf_symbol_new}--\eqref{eq:parent_symbol_new}.

The chosen $U_t$ must be consistent with the assignment $\eta_t^+$ (i.e., only robots in $C_\phi(\eta_t^+)$ execute leaf-relevant skills), and the edge cost is $\Delta T_t$ in~\eqref{eq:makespan_step_new}.

\textbf{Acceptance.}
Let $\varphi_{\mathrm{root}}$ denote the root node of $\mathcal{T}_h$.
\begin{equation}
\mathcal{F}\triangleq\{x_t\mid q_{\varphi_{\mathrm{root}}}(t)\in F_{\varphi_{\mathrm{root}}}\}.
\label{eq:accepting_set_new}
\end{equation}
We run on-the-fly Dijkstra to find a minimum-cost path to $\mathcal{F}$, minimizing
\begin{equation}
J(\pi)=\sum_{t=0}^{T-1}\Delta T_t.
\label{eq:objective_new}
\end{equation}

\subsection{Receding Horizon Execution and Replanning}\label{subsec:dynamic_execution}
The receding-horizon loop executes the first planned step, updates the formal state from perception, and replans the suffix; the three parts below cover progress, safety, and prediction.
Let $w_t$ be the perceived world state, $\ell_t\triangleq\ell(w_t)$ its discrete abstraction (workspace label over
$\{\mathcal{W}_j\}_{j=1}^{N_{\mathrm{ws}}}$), and $\mathrm{Proj}(w_t)$ the projection to the joint discrete robot state $\mathbf{s}_t$.

Define the planner initial node
\begin{equation}
x_t^0 \triangleq (\mathbf{s}_t,\eta_t^0,\mathbf{q}_t),\quad \mathbf{s}_t\leftarrow \mathrm{Proj}(w_t).
\label{eq:init_node_rhp}
\end{equation}

Here $\mathrm{Pre}(\mathbf{s}_t,\ell_t,a)$ is evaluated using the latest perception $w_t$ via the discrete abstraction and feasibility checks (e.g., reachability/collision).

\subsubsection{Plan Refinement}
At time $t$, solve
\begin{equation}
\pi_t^\star \triangleq \big\{(\eta_{t,k}^\star, U_{t,k}^\star)\big\}_{k=0}^{H_t-1}
\in \arg\min_{\pi\in\Pi(x_t^0\rightsquigarrow \mathcal{F})}\ \sum_{k=0}^{H_t-1}\max_{a\in U_{t,k}} d(a),
\label{eq:plan_opt_rhp}
\end{equation}
with $\mathrm{Pre}(\cdot)$ and $\sigma_\phi(\cdot)$ evaluated under current $\ell_t$.
Dispatch only $(\eta_{t,0}^\star,U_{t,0}^\star)$. Define progress event
$e_t=1 \Leftrightarrow$ all $a\in U_{t,0}^\star$ terminate successfully.
Upon $e_t$,
\begin{gather}
\mathbf{s}_{t^+}\leftarrow \mathrm{Proj}(w_{t^+}),\quad
\ell_{t^+}\leftarrow \ell(w_{t^+}), \notag\\
\mathbf{q}_{t^+}\leftarrow \mathrm{Update}(\mathbf{q}_t,\mathbf{s}_{t^+},\ell_{t^+},\eta_{t,0}^\star,U_{t,0}^\star).
\label{eq:update_state_monitor}
\end{gather}
Replan from $x_{t^+}^0=(\mathbf{s}_{t^+},\eta_{t^+}^0,\mathbf{q}_{t^+})$:
\begin{equation}
\pi_{t^+}^\star \in \arg\min_{\pi\in\Pi(x_{t^+}^0\rightsquigarrow \mathcal{F})} J(\pi).
\label{eq:suffix_replan}
\end{equation}

\subsubsection{Reactive Safety Constraints}
We define a workspace index map $\mathrm{ws}(a)\in\{1,\dots,N_{\mathrm{ws}}\}$ for skill instances and the active workspace set
$\mathcal{W}^{\mathrm{act}}(U)\triangleq \bigcup_{a\in U}\mathcal{W}_{\mathrm{ws}(a)}$.

Halt and replan if (i) \textbf{kinematic infeasibility} occurs, i.e., $\exists a\in U_{t,0}^\star$ with
$\mathrm{Pre}(\mathbf{s}_t,\ell_t,a)$ violated under latest $w_t$, or (ii) \textbf{intrusion risk} occurs:
the human is predicted to enter $\mathcal{W}^{\mathrm{act}}(U_{t,0}^\star)$ during a high-velocity maneuver.
In both cases preserve $\mathbf{q}_t$, update $\mathbf{s}_t\!\leftarrow\!\mathrm{Proj}(w_t)$ and $\ell_t\!\leftarrow\!\ell(w_t)$, then resolve~\eqref{eq:plan_opt_rhp}.

\subsubsection{Predictive Horizon Adaptation}
Given predicted trajectories $\hat{\vx}^{(i)}_{t+1:t+\Tpred}$, infer
\begin{equation}
\widehat{\mathrm{ws}}_{t+1}\triangleq
\arg\max_j\ \sum_{\tau=1}^{\Tpred}\mathbb{I}\big[\hat{\vx}^{(i)}_{t+\tau}\in\mathcal{W}_j\big].
\label{eq:ws_predict_new}
\end{equation}
Construct $w_{\mathrm{pred}}$ by replacing only the human-workspace attribute in $w_t$ with $\widehat{\mathrm{ws}}_{t+1}$ and solve in parallel
\begin{gather}
x_{\mathrm{pred}}^0\triangleq(\mathrm{Proj}(w_{\mathrm{pred}}),\eta_t^0,\mathbf{q}_t), \notag\\
\pi_{\mathrm{pred}}^\star\in\arg\min_{\pi\in\Pi(x_{\mathrm{pred}}^0\rightsquigarrow \mathcal{F})} J(\pi).
\label{eq:plan_pred_new}
\end{gather}
Take over only when $\ell(w_t)=\ell(w_{\mathrm{pred}})$:
\begin{equation}
\textsc{Takeover}\ \Leftrightarrow\ \ell(w_t)=\ell(w_{\mathrm{pred}}),\quad \pi_t\leftarrow \pi_{\mathrm{pred}}^\star.
\label{eq:takeover_new}
\end{equation}

\begin{figure}[tbp]
\centering
\includegraphics[width=0.9\linewidth]{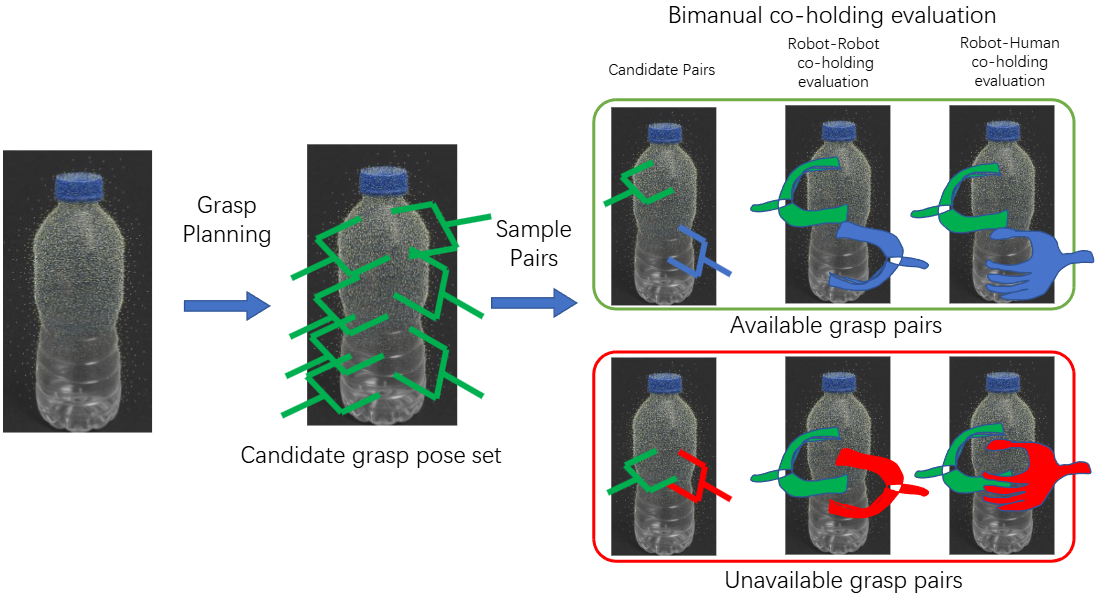}
\caption{Pick planning pipeline. }
\label{fig:grasp_eval}
\vspace{-5pt}
\end{figure}
\section{R2R \& R2H Handover Planning}
\label{sec:4}
To accomplish collaborative handover tasks with multiple manipulators, we define four general-purpose skills that cover diverse scenarios: pick, handover\_r2r, handover\_r2h, and place. To balance user experience and operational efficiency, motion planning for all skills runs in a planning mode whenever possible. Instead of targeting a single deterministic pose \cite{cbs}, each skill considers a set of feasible goal states. Multiple candidate trajectories are planned in parallel, enabling fast solutions that reduce user waiting time while improving efficiency.
\subsection{Pick and Place}
We focus on cylindrical objects. We generate a library of grasp candidates based on object point clouds. In collaborative handover tasks, it is essential to ensure that the receiving side has sufficient space to grasp the object. We filter these candidates by simulating a co-grasping scenario with a human hand model, discarding pairs that result in collision or are too close, as shown in Fig.~\ref{fig:grasp_eval}. The remaining feasible pairs form the target set for motion planning. For each pair of grasps, the desired grasp of the receiving arm is defined as the complementary configuration of the pair. 

When manipulators hold different objects and a direct handover cannot be performed, we define the place skill. For placement, valid poses are sampled from planar surfaces within the intersection of the robots' reachable regions.

\begin{figure}[tbp]
\centering
\includegraphics[width=\linewidth]{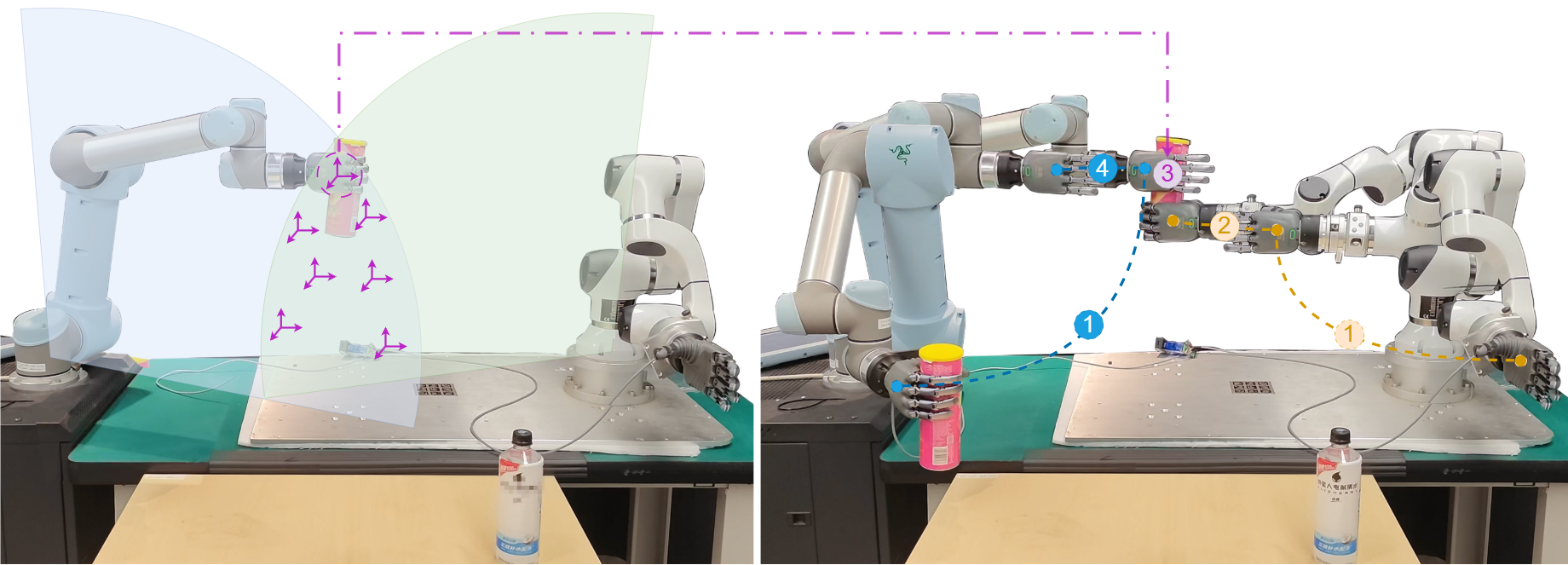}
\caption{The proposed robot-to-robot handover method. }
\label{fig:handover_r2r}
\vspace{-5pt}
\end{figure}

\subsection{Robot-to-robot Handover}
 We define the robot-to-robot (R2R) handover (handover\_r2r)  skill, which enables inter-robot handovers to extend the reach of objects. Since there exist infinitely many candidate poses within the overlap of two robots’ workspaces, we optimise the handover position using two criteria:

Feasibility -- Each robot's reachable region is modelled as a sphere derived from its base position and reach radius. We identify the inscribed sphere at the intersection of the two reachable regions and uniformly sample candidate poses within it. Feasibility increases as the pose approaches the spherical intersection centre.

Time efficiency -- For each sampled pose, we compute the distance from both manipulators' current end-effectors and take the maximum distance as a proxy for execution time. In addition, optimal handover orientations are calculated by considering the current joint states of both arms and candidate grasp poses.

Once a feasible handover pose is determined, we address the collision-avoidance challenge inherent to contact states. As shown in Fig.\ref{fig:handover_r2r}, the receiving arm first moves to a pre-handover position that is collision-free, then executes a linear motion along the approaching vector to achieve secure grasping. After the transfer, the giving arm retreats by a short distance to reduce the risk of collisions in subsequent motions.

\subsection{Robot-to-Human Handover}
Unlike R2R handover, R2H prioritizes user ergonomics over kinematic optimality. We define the handover target within a comfort zone (forearm raised 90°–120° close to the torso), allowing the planner to adapt to the user's estimated posture.

\begin{figure}[tbp]
\centering
\includegraphics[width=\linewidth]{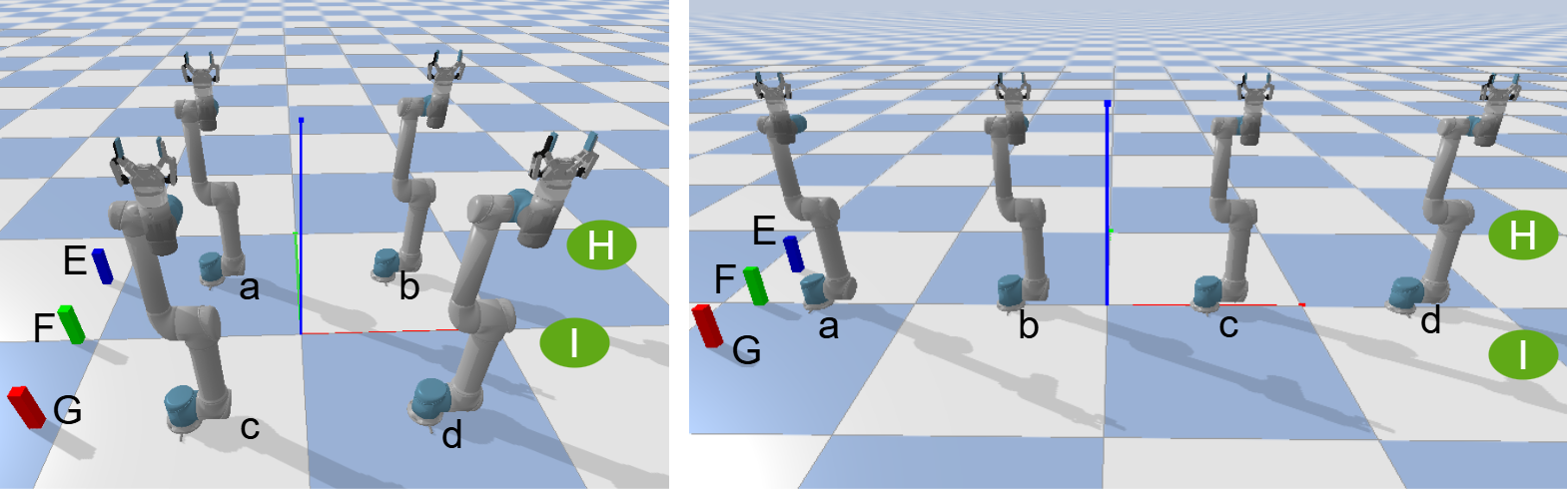}
\caption{Simulation experiment setup for 2-4-square and 2-4-line.\protect\footnotemark}
\label{fig:sim_setup}
\vspace{-5pt}
\end{figure}

\section{\textsc{Experimental Results and Evaluations}}
In our implementation, we employ \texttt{GPT-4o-mini} as the reasoning engine, balancing semantic understanding with the rapid inference speed necessary for our receding horizon framework.
In this section, we design a series of experiments to address the following research questions:

\textbf{Q1:} Can multiple manipulators collaboratively extend their combined workspace to serve users with object handovers over an extensive spatial range?

\textbf{Q2:} Can the proposed dynamic planning mechanism effectively handle dynamic scenarios, such as moving users or dynamically changing task instructions?

\textbf{Q3:} Can the H-LTL$_{f}$ planning module improve efficiency when processing multiple simultaneous tasks?

\textbf{Q4:} Can the framework effectively avoid/resolve conflicts during collaborative handovers?

\subsection{Multi-robot collaborative handover in dynamic environments}
To evaluate the scalability and robustness of the proposed approach, we developed a custom simulation environment that simulates unforeseen environmental dynamics, specifically \textbf{(1)} real-time changes in receiver locations and \textbf{(2)} the injection of new tasks during execution. We procedurally generate long-horizon collaborative tasks involving varying numbers of objects and logical constraints. The complexity is categorised by the number of base tasks ($N_{\text{task}} \in \{1, 2, 4\}$), the number of available fixed-base robots ($N_{\text{robot}} \in \{1, 2, 4\}$), and the topological layout of fixed-base robots (square vs.\ line), resulting in a test suite ranging from simple single-robot executions (1-1) to complex multi-robot coordination scenarios (4-4-square, 4-4-line), as shown in Fig.~\ref{fig:sim_setup}.

\footnotetext{This figure illustrates the setup for the simulation experiments and does not represent the sim setup for the real-world hardware experiment.}

Following the baseline used in \cite{xu2024nl2hltl2plan,wang2025dartllmdependencyawaremultirobottask}, we compare our method with SMART-LLM \cite{kannan2023smart}, an LLM-based multi-robot task planner that generates Python scripts with predefined action APIs for task decomposition and task allocation. We implement an enhanced version, \cite{kannan2023smart}-R, that replans from the updated simulator state when a new task arrives or the next action violates reachability, possession, or collision preconditions. To ensure a fair comparison, SMART-LLM-R is evaluated under the same action API, world-state update pipeline, and execution-level feasibility checks, including reachability, object possession, and collision constraints. We focus our quantitative comparison on this baseline because it can be instantiated under the same executable action interface, whereas other related systems in Table~\ref{tab:methods_comparison} differ in task assumptions, skill libraries, state abstractions, and deployment settings, making a direct quantitative comparison less controlled. We evaluate performance using three key metrics:
\noindent\textbf{(i) Success Rate}: The percentage of trials where all goal conditions are met without violation of kinematic feasibility.
\noindent\textbf{(ii) Time Cost}: The total time required to finish all tasks (in seconds), reflecting the efficiency of task allocation and motion planning.
\noindent\textbf{(iii) Token Usage}: The average number of tokens consumed per query to the LLM, representing the computational cost and latency of the reasoning module.

The statistical results are presented in Table~\ref{tab:sim_results}. Our framework demonstrates superior performance across all dimensions, particularly as problem complexity increases.
\noindent\textbf{Scalability (Q1)}: While SMART-LLM performs adequately on simple tasks (1-1, 1-2), its success rate degrades significantly as the number of tasks increases (dropping to 19\% in the 4-4-line setting). This is largely because generating long, monolithic Python scripts increases the probability of logical inconsistencies and syntax errors. In contrast, our method maintains a high success rate (93\%) even in the most complex scenarios, validating the robustness of the H-LTL$_f$ structure.\\
\noindent\textbf{Plan Quality (Q3)}: Our method achieves lower time costs compared to the baseline in most settings. The formal planning engine optimises task allocation globally based on the unified search graph, whereas SMART-LLM relies on the LLM's inherent (and often suboptimal) sequencing bias.\\
\noindent\textbf{Token Efficiency}: A distinct advantage of our approach is the significant reduction in token usage. By extracting concise hierarchical specifications rather than generating verbose executable code, our method reduces token consumption by 80-94\%. This efficiency is critical for real-time dynamic replanning, enabling faster query responses and lower operational costs.
\begin{table}[t]
\centering
\caption{Performance comparison in simulation. Metrics reported are Success Rate (\%), Time Cost (s), and Token Usage.}
\vspace{-10pt}
\label{tab:sim_results}
\resizebox{\linewidth}{!}{%
\begin{tabular}{ccccccc}
\toprule
\multirow{2}{*}{\makecell{Tasks–Robots}} 
    & \multicolumn{2}{c}{\makecell{Success Rate(\%) $\uparrow$}} 
    & \multicolumn{2}{c}{\makecell{Time Cost(s)$\downarrow$}} 
    & \multicolumn{2}{c}{\makecell{Token Usage  $\downarrow$}} \\
\cmidrule(lr){2-3} \cmidrule(lr){4-5} \cmidrule(lr){6-7}
& Ours & \cite{kannan2023smart}-R & Ours & \cite{kannan2023smart}-R & Ours & \cite{kannan2023smart}-R \\
\midrule
1-1 & \textbf{100} & 100 & \textbf{1.0} & 1.0 & \textbf{261} & 2690 \\
2-1 & \textbf{100} & 99 & \textbf{2.1} & 2.1 & \textbf{470} & 3252\\
4-1 & \textbf{100} & 96 & \textbf{4.4} & 4.4 & \textbf{885} & 4493 \\
\midrule
1-2 & \textbf{100} & 100 & \textbf{2.1} & 2.1 & \textbf{264} & 3611 \\
2-2 & \textbf{100} & 68 & \textbf{2.2} & 2.8 & \textbf{473} & 4386 \\
4-2 & \textbf{97} & 41 & \textbf{4.5} & 6.7 & \textbf{896} & 6595 \\
\midrule
1-4-line & \textbf{100} & 79 & \textbf{4.4} & 4.4 & \textbf{374} & 4616 \\
2-4-line & \textbf{99} & 39 & \textbf{9.5} & 9.5 & \textbf{607} & 4907 \\
4-4-line & \textbf{93} & 19 & 19.4 & \textbf{19.3} & \textbf{1259} & 7385 \\
\midrule
1-4-square & \textbf{100} & 94 & \textbf{2.1} & 2.1 & \textbf{264} & 4243 \\
2-4-square & \textbf{100} & 40 & \textbf{2.3} & 2.7 & \textbf{471} & 5598 \\
4-4-square & \textbf{95} & 22 & \textbf{4.7} & 7.9 & \textbf{902} & 7587 \\
\bottomrule
\end{tabular}
}
\vspace{-5pt}
\end{table}

\subsection{Real-world multi-robot collaboration involving humans}
\begin{figure}[tbp]
\centering
\includegraphics[width=\linewidth]{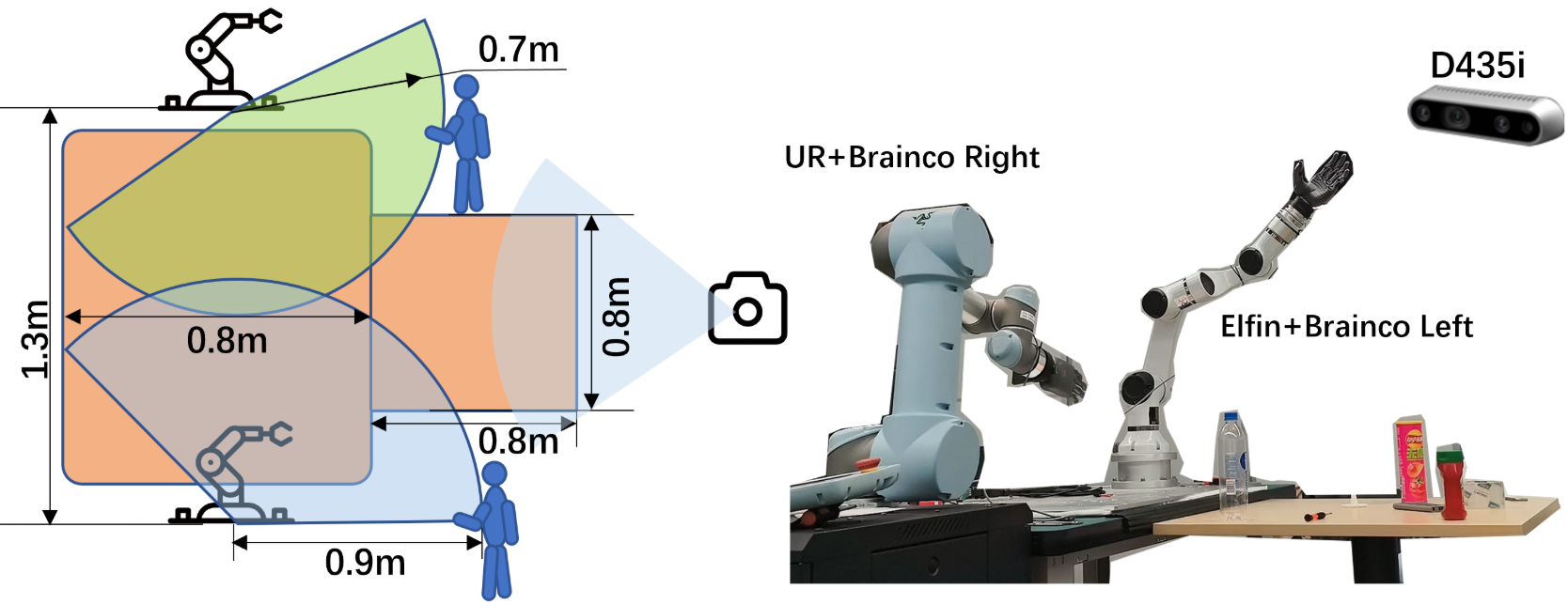}
\caption{Real-world experiment setup.}
\label{fig:experiment_setup}
\vspace{-5pt}
\end{figure}

To demonstrate real-world feasibility, as illustrated in Fig.~\ref{fig:experiment_setup}, our experimental setup consists of two heterogeneous manipulators, namely an Elfin-3 and a UR5, which collaborate to perform handover tasks. The manipulators are equipped with the left and right BrainCo robotic hands, respectively, with effective reaching radii of 0.7 m and 0.9 m. An Intel RealSense D435 camera captures both RGB and depth images of the environment.

We construct a dataset containing six experimental scenarios, where nearby objects are in the same robot's workspace as the user, and distant objects require R2R handover for delivery:

\textbf{T1:} A single user requests a distant object  (Q1).

\textbf{T2:} A single user requests a distant object while walking into the other robot's workspace during the handover process (Q2).

\textbf{T3:} A single user requests a nearby object and, after a short delay, issues a new request for another distant object (Q2).

\textbf{T4:} Two users located in different robots' workspaces simultaneously request nearby objects; then a new request is issued to deliver an object in the shared workspace to either user (Q3).

\textbf{T5:} Two users located at different robots' workspaces simultaneously request distant objects that both require inter-robot collaboration (Q4).

\textbf{T6:} Two users located at different robots' workspaces request nearby objects while switching their positions during the task, leading to combined challenges of dynamic adaptation and multi-task coordination (Q2, Q3, Q4).

\begin{figure}[tbp]
\centering
\includegraphics[width=\linewidth]{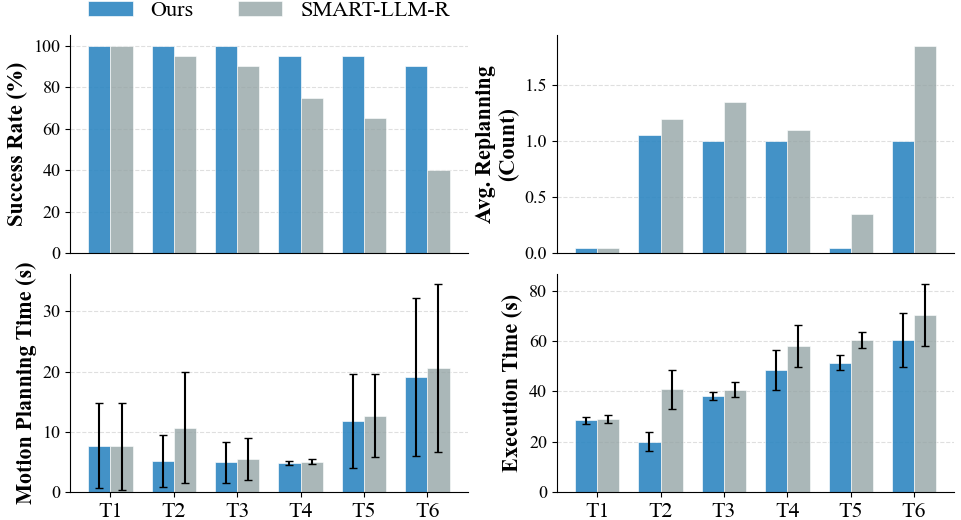}
\caption{Statistical results from real-world experiments. Avg. Replanning Count denotes the average number of plan regenerations required to complete a task (trials exceeding 3 are marked as failures).}
\label{fig:exp_result}
\end{figure}
We first evaluate whether the proposed system can sustain robust performance when users move during execution. For each task, we conducted 20 experimental trials. As summarised in Fig.~\ref{fig:exp_result}, our method maintains better performance across all tasks (T1--T6). In the dynamic interruption scenario (T2), our system reduces execution time by over 50\%. This validates our reactive safety constraints: the system detects the kinematic infeasibility of the planned handover caused by user movement, triggering an immediate halt and efficient recovery. In complex conflict scenarios (T5, T6), our method maintains a significantly lower replanning count compared to the baseline. This confirms that the unified search graph effectively prevents the generation of invalid plans or deadlocks that force the baseline into frequent iterative retries. These results indicate that our H-LTL$_f$-based receding horizon planning pipeline sustains both correctness and responsiveness under real-time human motion.

\subsection{Advantages of Receding Horizon Formal Planning over Pure LLM Generation}

This subsection evaluates Secs.~\ref{subsec:team_models} and~\ref{subsec:dynamic_execution}. While LLMs exhibit remarkable flexibility in decomposing natural language into action sequences, they function as probabilistic token generators and lack an internal world model, often producing plans that are logically invalid or kinematically infeasible in multi-robot settings. By grounding LLM outputs into a receding horizon formal planning engine based on a unified search graph, our framework provides two critical advantages:

\noindent\textbf{(i) Dynamic Efficiency via Receding Horizon Optimization.}
While LLMs struggle to optimise schedules under stochastic human interactions, our framework ensures execution efficiency through the horizon update and plan refinement mechanism described in Sec.~\ref{subsec:dynamic_execution}. Unlike static planners that rely on estimated durations, our system monitors state transitions and resolves the unified search graph when sub-tasks are completed. For instance, in T4, if a robot finishes a handover early, the RHP loop immediately generates a new optimal suffix path to assign pending tasks to the free agent, reducing team idle time rather than waiting for a rigid schedule, as reflected in Fig.~\ref{fig:exp_result}.

\noindent\textbf{(ii) Correct-by-Construction Safety and Feasibility.} 
 The formal planning engine operates on a strict transition system (the Product Team Model). It enforces preconditions and mutual-exclusion constraints encoded in the automaton, rejecting action sequences that violate the system dynamics. Two typical failure modes inherent to pure LLM planning, which our formal engine prevents, are illustrated in Fig.~\ref{fig:llm_failure_cases}:
\begin{figure}[htbp]
\centering
\includegraphics[width=\linewidth]{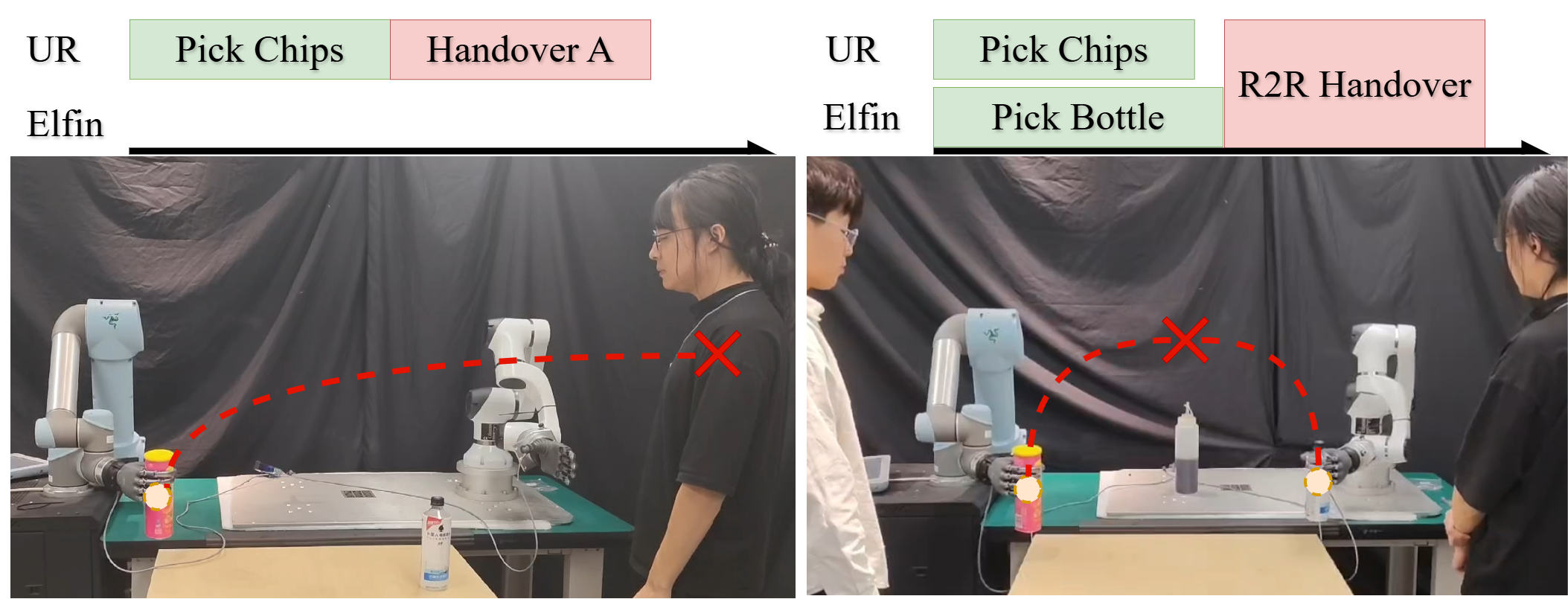}
\caption{Typical LLM failure modes prevented by formal 
planning checks.}
\label{fig:llm_failure_cases}
\vspace{-0.3cm}
\end{figure}
\textit{Shared Object Conflict:} The LLM occasionally proposes a ``simultaneous-hold'' handover, effectively assuming that robots can rigidly grasp multiple objects without a synchronised transfer protocol. Our formal team model explicitly encodes mutual exclusion regarding object possession and instead generates a safe two-phase exchange (placing followed by a regrasp). 

\textit{Kinematic Violation:} To minimise sequence length, the LLM often commands a robot to \texttt{pick/handover} an object outside its physical workspace. The formal planner invalidates any plan containing actions that are unreachable in the robot's transition graph and searches for a viable alternative path.

\subsection{Reducing Replanning Latency via Predictive Planning and Parallel Motion Planning}
This ablation evaluates Sec.~\ref{subsec:dynamic_execution} and Sec.~\ref{sec:4}. Operating in human-centric, time-sensitive settings requires minimising perceived latency. Two bottlenecks are prominent: (i) LLM-integrated high-level replanning, and (ii) motion planning when multiple manipulators must coordinate (e.g., R2R handover). We address both by: 
\textbf{(a) Predictive planning}, which forecasts short-horizon human trajectories and preemptively submits a replanning request when reachability assumptions are likely to change; and 
\textbf{(b) Parallel motion planning} for multi-robot skills (e.g. R2R handover), which spawns candidate trajectories simultaneously for different arms and grasp pairs.
\begin{table}[htbp]
\vspace{-10pt}
\centering
\small
\caption{\textbf{Ablation on HRC fluency metrics over all tasks}. }
\vspace{-10pt}
\resizebox{0.45\textwidth}{!}{%
\begin{tabular}{lcccc}
\toprule
Variant & \makecell{H-IDLE\\$\downarrow$ (\%)} & \makecell{R-IDLE\\$\downarrow$ (\%)} & \makecell{C-ACT\\$\uparrow$ (\%)} & \makecell{F-DEL\\$\downarrow$ (\%)}\\
\midrule
\textbf{Ours (full)} & $\mathbf{67.1\pm17.2}$ & $\mathbf{38.8\pm10.1}$ & $\mathbf{64.9\pm11.9}$ & $\mathbf{-4.5\pm8.6}$ \\
w/o P.P.  & $70.8\pm16.7$ & $45.4\pm11.3$ & $63.5\pm15.0$ & $-4.4\pm7.9$ \\
w/o P.M.P. & $68.6\pm17.9$ & $42.7\pm11.9$ & $60.8\pm14.0$ & $-4.2\pm8.6$ \\
\bottomrule
\end{tabular}%
}
\parbox{\linewidth}\small{Abbrev.: \textbf{P.P.} = Predictive planning, \textbf{P.M.P.} = Parallel motion planning}
\vspace{-10pt}
\label{tab:hrc}
\end{table}

Following the HRC fluency metrics in \cite{garcia2020human}, we report Human Idle Time (H\textendash IDLE), Robot Idle Time (R\textendash IDLE), Concurrent Activity (C\textendash ACT), and Functional Delay (F\textendash DEL).\footnote{H\textendash IDLE/R\textendash IDLE: percentage of total task time the respective agent is inactive; C\textendash ACT: percentage of time at least two agents are concurrently active; F\textendash DEL: percentage of time between one robot finishing and the other starting. Higher C\textendash ACT is better; lower H\textendash IDLE, R\textendash IDLE, F\textendash DEL are better.} 
We conduct an ablation across all tasks, comparing our complete system against variants that (i) remove the predictive trigger and (ii) remove parallel planning. Results are shown in Table~\ref{tab:hrc}. Both components contribute: predictive planning chiefly reduces human and robot waiting (H\textendash IDLE, R\textendash IDLE), while parallel planning primarily cuts robot waiting (R\textendash IDLE) and improves synchrony (C\textendash ACT).
\section{Conclusion}
In this work, we propose a neuro-symbolic framework that grounds LLM outputs into H-LTL$_f$ for dynamic, human-aware multi-robot planning, and extends H-LTL$_f$ from single-robot leaf execution to strongly-coupled cooperative STAP via a coalition-aware unified search graph. Experiments in simulation and on real robots demonstrate higher success rates, better interaction fluency, and lower replanning latency than baselines. 
\bibliographystyle{ieeetr}
\bibliography{ref}
\end{CJK}
\end{document}